\documentclass[10pt,twocolumn,letterpaper]{article}

\usepackage[arxiv]{wacv} 

\usepackage{graphicx}
\usepackage{amsmath}
\usepackage{amssymb}
\usepackage{booktabs}
\usepackage{multirow}

\usepackage[pagebackref,breaklinks,colorlinks]{hyperref}

\usepackage[capitalize]{cleveref}
\crefname{section}{Sec.}{Secs.}
\Crefname{section}{Section}{Sections}
\Crefname{table}{Table}{Tables}
\crefname{table}{Tab.}{Tabs.}
\usepackage{soul}

\def\wacvPaperID{1328} 
\def\confName{WACV}
\def\confYear{2025}
\begin{document}

\title{EmoVOCA: Speech-Driven Emotional 3D Talking Heads}

\author{Federico Nocentini\\
University of Florence, Italy\\
{\tt\small federico.nocentini@unifi.it}
\and
Claudio Ferrari\\
University of Parma, Italy\\
{\tt\small claudio.ferrari@unipr.it}
\and
Stefano Berretti\\
University of Florence, Italy\\
{\tt\small stefano.berretti@unifi.it}
}
\twocolumn[{
\maketitle
\vspace{-1.2cm}
\begin{center}
\includegraphics[width=\textwidth]{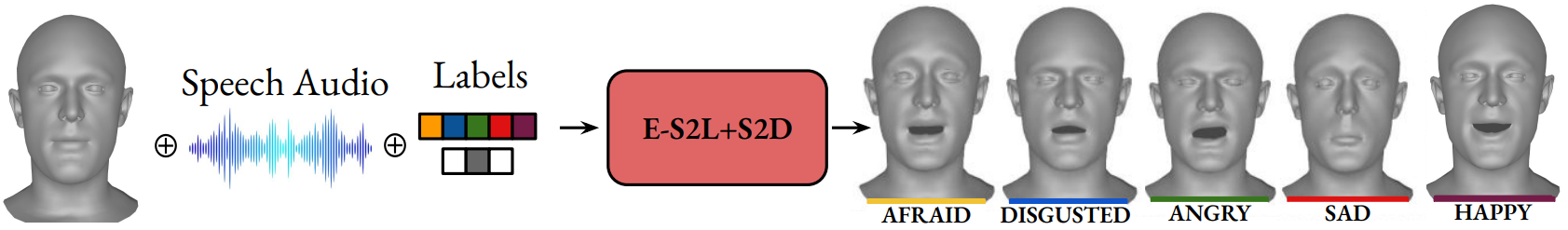}
\captionof{figure}{We introduce \textbf{EmoVOCA}, a novel approach for generating a synthetic 3D Emotional Talking Heads dataset which leverages speech tracks, intensity labels, emotion labels, and actor specifications. The proposed dataset can be used to surpass the lack of 3D datasets of expressive speech, and train more accurate emotional 3D talking head generators as compared to methods relying on 2D data as proxy.}
\label{fig:core}
\end{center}
}]


\begin{abstract}
A notable challenge in 3D talking head generation consists in blending speech-related motions with expression dynamics. This is primarily caused by the lack of comprehensive 3D datasets that combine diversity in spoken sentences with a variety of facial expressions. Some literature works attempted to overcome such lack of data by fitting parametric 3D models (3DMMs) to 2D videos, and using the reconstructed 3D faces as replacement. However, 
their underlying parametric space limits the precision required to accurately 
reproduce convincing lip motions 
and synching, which is crucial for the application at hand. In this work, we look at the problem from a different perspective, and developed a data-driven technique to combine inexpressive 3D talking heads with a set of 3D expressive sequences, which we used for creating a synthetic dataset, called \textbf{EmoVOCA}. We then designed and trained an emotional 3D talking head generator that accepts a 3D face, an audio file, an emotion label, and an intensity value as inputs, and learns to animate the audio-synchronized lip movements with expressive traits of the face. Comprehensive experiments, both quantitative and qualitative, using our data and generator evidence superior ability in synthesizing convincing animations, when compared with the best performing methods in the literature. Our code and pre-trained models are available at \href{https://github.com/miccunifi/EmoVOCA}{https://github.com/miccunifi/EmoVOCA}.
\vspace{-0.5cm}
\end{abstract}

\section{Introduction}
\label{sec:intro}
%
Generating 3D talking heads from speech 
aims at animating a 3D facial model with dynamic lip movements that correspond to a spoken sentence. While current state-of-the-art methodologies such as~\cite{nocentini2023learning, FaceDiffuser_Stan_MIG2023, fan2022faceformer, xing2023codetalker, peng2023selftalk, emote, peng2023emotalk, nocentini2024scantalk3dtalkingheads, PAIER2023101199, nocentini2024fixedtopologiesunregisteredtraining} effectively replicate lip-syncing and facial deformations, achieving natural emotional expression in animated faces lacks full conviction. 
The primary constraint inhibiting research progress in emotional 3D talking heads resides in the absence of suitable 3D datasets that allow learning the complex interaction between geometrical mouth deformations induced by speech and expressions. 
In fact, directly collecting such data is a rather unpractical solution due to costly devices and a time-consuming capturing process. 
A workaround that was explored recently consists in leveraging 2D video datasets as a proxy. For example, Peng~\etal~\cite{peng2023emotalk} and Daněček~\etal~\cite{emote} extracted per-frame 3D faces from video datasets of emotional speech thanks to parametric 3D models (3DMMs). In doing so, they gathered sequences of plausible 3D faces paired with the speech extracted from the videos. A shortcoming of these approaches, though, is that the precision and variety of the 3D estimated facial deformations are bounded by the underlying parametric model, with a consequent loss of subtle lip movements. In fact, while 3DMMs are good for modeling facial shapes and expressions, they are not sufficiently expressive for reproducing speech-related lip motions. 

In the attempt of solving the above, we explore a different approach and propose to combine an existing 3D talking head dataset, where captured subjects only show a neutral emotional state (VOCAset~\cite{Cudeiro_2019_CVPR}), with 3D emotional faces from a dataset of 3D expressive sequences (Florence4D~\cite{10042606}). We chose these specific datasets as the meshes they include share the same topology. The major challenge here consists in realistically combining speech-related mouth motions with expression-induced facial deformations. To simplify the problem, some methods~\cite{richard2021meshtalk} consider an additional constraint, and assume that facial deformations resulting from expressions are isolated in the upper half of the face, forcing a spatial disentanglement of such motions. While this solution can work for natural facial movements, the above assumption only partially holds true when emotions are involved. Suffice it to say, anger or happiness induce changes in the mouth shape, which in turn influence its motion during speech. 
We here tackle the challenge of combining speech and expression induced facial deformations from 3D face datasets and surpass the need for parametric 3D models, in order to build training data for emotion-conditioned 3D talking head generators. To this aim, we propose a framework composed of two encoders and a single, shared decoder.
The idea is that each encoder learns speech or expression specific features, while the decoder learns both. Once trained, we can combine speech and expression related motions by simple feature combination, and let the decoder generate the entangled deformations. 
To showcase the applicability of our synthesized data, we used them to train a network that generates emotional 3D talking heads from a speech track, an intensity label, an emotion label, and an actor to animate, as illustrated in~\cref{fig:core}. To this end, we developed on two state-of-the-art solutions~\cite{fan2022faceformer, nocentini2023learning} that generate unexpressive talking heads based on the sole audio features, and adjusted them to be conditioned with an emotion and an intensity label. In summary the main contributions of our work are:
\begin{itemize}
    \item We introduce a new data-driven approach to realistically \textit{entangle} speech and expression specific facial deformations, providing more accurate training data for building expressive talking head generators;
    \item With the above, we synthesize EmoVOCA, a new 3D dataset combining speech with emotional nuances;  
    \item On such data, drawing from state-of-the-art techniques, we designed and trained two deep architectures that generate accurate expressive 3D talking heads using an audio, an emotion, and intensity labels;
    \item In an extensive set of experiments, we unveil limitations of previous methods, and the advantages of our scheme with respect to prior works.  
\end{itemize}

\section{Related Work}\label{sec:related}
First efforts primarily concentrated on the synthesis of facial animation through the manipulation of pre-defined facial rigs using procedural mechanisms. 
Subsequent to this, facial animations were constructed through viseme-dependent co-articulation models or via the blending of facial templates~\cite{DEMARTINO2006971, 10.1145/2897824.2925984, 982373}. These methods showcased a synthesis framework that combines language and facial expressions at a fundamental level.
Karras~\etal~\cite{10.1145/3072959.3073658} harnessed a dataset-driven approach, learning a 3D facial animation model from a limited but high-quality 3D dataset of a specific actor. This strategy stands in contrast to VOCA~\cite{Cudeiro_2019_CVPR}, which leveraged a more expansive dataset featuring diverse subjects, capable of animating an array of corresponding identities from audio cues. 
\textit{MeshTalk}~\cite{richard2021meshtalk} proposed a complementary solution to those discussed above by learning a categorical representation for facial expressions. Sampling in an auto-regressive way from this categorical space, this approach can animate a given 3D facial template mesh of a subject from audio inputs. 
\textit{FaceFormer}~\cite{fan2022faceformer} proposed a Transformer-based autoregressive model, which encodes the long-term audio context and autoregressively predicts a sequence of animated 3D face meshes.
Nocentini~\etal~\cite{nocentini2023learning}, aimed to address the challenges encountered by FaceFormer and VOCA, like the long training time, and increased the lip-movements accuracy, introducing S2L+S2D. This framework enhances the lip-sync abilities of talking heads by guiding them through a landmark-based motion paradigm. 
In the most recent effort to develop speech-driven 3D talking heads, Stan~\etal proposed \textit{FaceDiffuser}~\cite{FaceDiffuser_Stan_MIG2023}, an approach centered around a Gated Recurrent Unit model. This model was trained to operate like a diffusion model, predicting the 3D face from a set of Gaussian noise inputs. 
We also mention the work by Thambiraja~\etal~\cite{Thambiraja_2023_ICCV}, who proposed \textit{Imitator} to add personalized traits to the talking head.

All the aforementioned methods were trained on publicly available datasets. However, these datasets pose a limitation as the recorded faces lack expressions, resulting in a deficiency of emotional content in the generated faces.
To address the lack of specific data, researchers have tested various techniques. Lu~\etal~\cite{lu2023audiodriven} proposed to use EMOCAv2~\cite{danvevcek2022emoca} to reconstruct 3D talking heads from in-the-wild videos. 
Chang~\etal~\cite{https://doi.org/10.1002/cav.2076} proposed a fusion of 2D and 3D datasets to inject emotional information into the generated faces. 
Conversely, Peng~\etal~\cite{peng2023emotalk} devised a methodology to reconstruct 3D emotional talking head datasets from 2D sources based on a facial blendshape capturing method. Upon constructing the dataset, they developed a Transformer-based model, called \textit{Emotalk}, for generating 3D emotional talking heads. Differently from others, they did not condition the generation with emotion labels, but directly extracted emotional context from the audio. 
More recently, Sung-Bin~\etal~\cite{sungbin2023laughtalk} proposed LaughTalk, a framework proficient in crafting speech-driven 3D laughing talking heads. Additionally, Daněček~\etal~\cite{emote} presented EMOTE, a framework capable of generating emotional 3D talking heads predicated on audio input and emotion labels. 

Again, the very core of these approaches is based upon the efficacy of reconstructing a 3D face from 2D representations. We will show that, 
this strategy results in decreased lip-sync accuracy. 

\vspace{-0.3cm}
\section{Proposed Approach}\label{sec:proposed}
The lack of publicly available emotional 3D talking head datasets motivated our efforts. 
We propose here a solution to address most of the limitations faced by previous works that attempted to exploit 2D video data as an intermediate means to gather 3D expressive talking heads. To this aim, we leverage two publicly available 3D face datasets: one includes 3D sequences of talking faces in \textit{neutral expression}; the other, instead, comprises sequences of 3D faces portraying \textit{emotion related facial expressions}, yet without speech-related lip movements.
No additional assumptions are imposed on the data; for instance, sequences are not required to have uniform lengths, nor is synchronization necessary between instances in the two datasets. 

\begin{figure*}[!ht]
    \centering
    \begin{subfigure}[b]{0.49\linewidth}
        \centering
        \includegraphics[width=\textwidth]{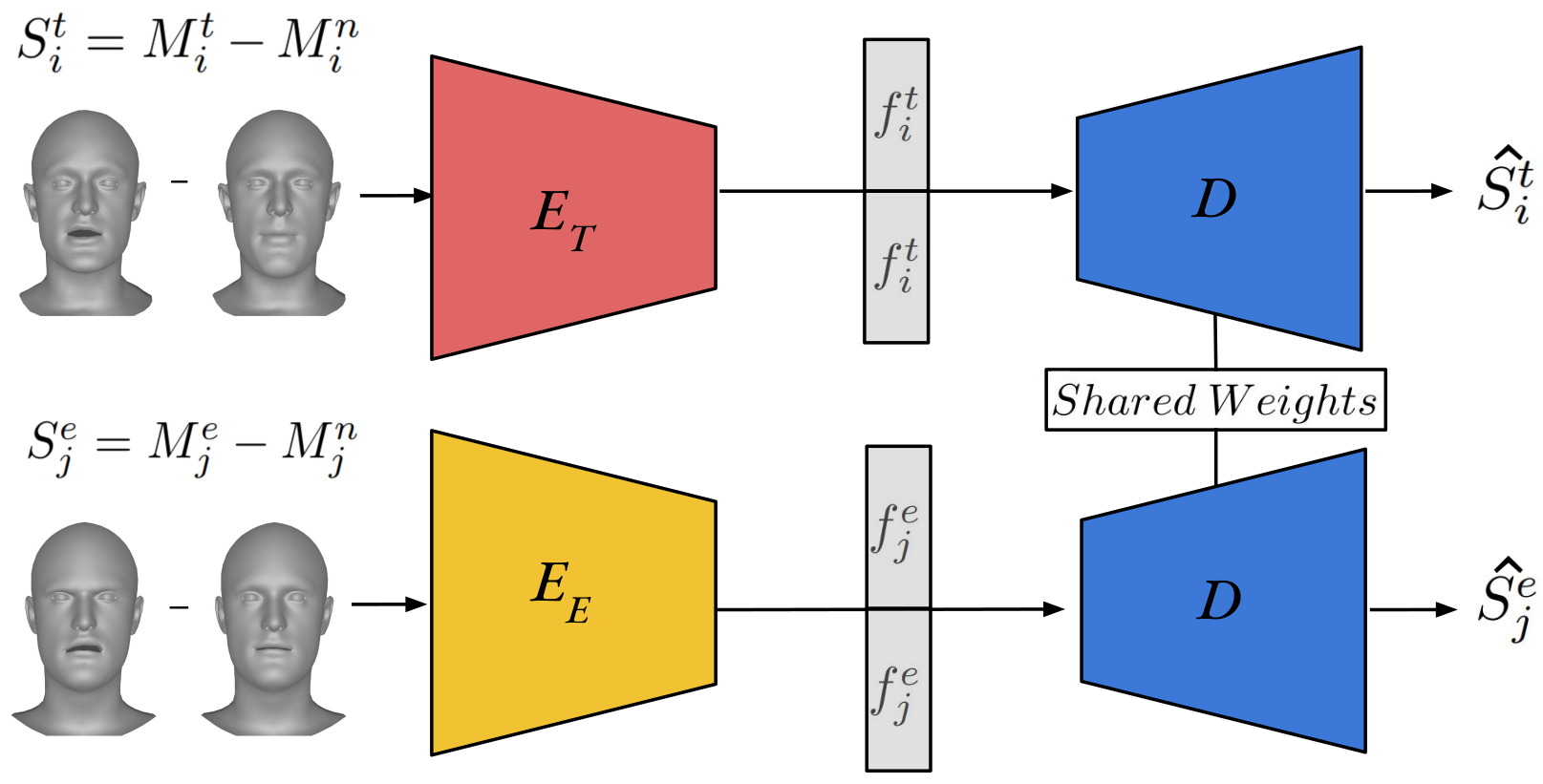}
        \caption{\textbf{Training phase} of our model: In this stage, $E_T$ is utilized when the input is sourced from the $S_T$ dataset, and $E_E$ is employed when the input is derived from the $S_E$ dataset.}
        \label{fig:training_phase}
    \end{subfigure}%
    \hfill
    \vline
    \hfill
    \begin{subfigure}[b]{0.49\linewidth}
        \centering
        \includegraphics[width=\textwidth]{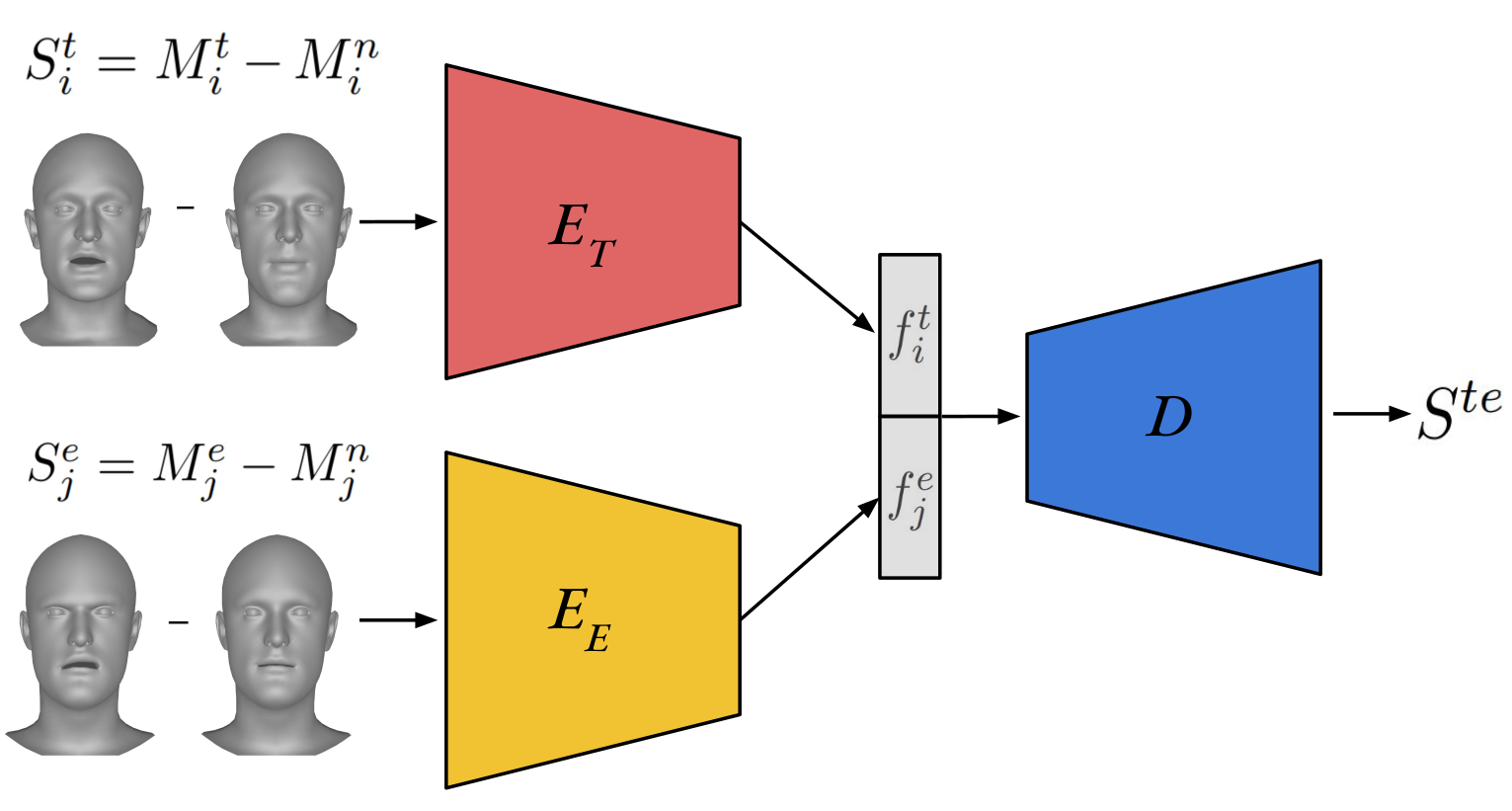}
        \caption{\textbf{Inference phase} of our model: During this phase, the features extracted from both encoders are combined through concatenation and subsequently forwarded to the decoder.}
        \label{fig:inference_phase}
    \end{subfigure}
    \caption{Overview of our framework. (a) Two distinct encoders, namely $E_T$ and $E_E$ process the talking and expressive 3D head displacements, separately, while a common decoder $D$ is trained to reconstruct them. 
    (b) At inference, talking and emotional heads are combined by concatenating the encoded latent vectors, and the decoder outputs a combination of their displacements.}
    \vspace{-0.4cm}
    \label{fig:overview}
\end{figure*}

Our idea is to generate emotional 3D talking heads by learning to explicitly \textit{entangle} speech and emotion face deformations by mixing latent codes that are separately learned from the two datasets. The framework, illustrated in~\cref{fig:overview}, is composed of two encoders, which separately learn to embed speech or emotion related facial deformations, and a shared decoder. The model is trained to reconstruct the input deformations. Being the decoder shared across the two datasets, at inference time the embeddings can be combined so to generate the entangled deformations. 
Ultimately, this allowed us to create a new dataset, called EmoVOCA, that can be employed for training emotional 3D talking heads generation approaches. \\

\subsection{Data Preparation}\label{sec:data_preprocess}
We utilized two 3D face datasets, referred to as $D_T$ and $D_E$: $D_T$ contains \textit{inexpressive} 3D talking heads, while $D_E$ consists of \textit{emotional} 3D sequences without speech-driven lip activation. Both $D_T$ and $D_E$ are in FLAME~\cite{FLAME:SiggraphAsia2017} topology. This choice simplified the framework design by allowing us to use the same encoder architecture, but does not represent a strict constraint. The two encoders do not share weights and are trained separately (\cref{sec:framework}), so they can be adapted to process 3D faces of different topologies.
In order to learn speaker-independent speech or expression deformations, we first pre-process the data to remove the identity component from each face. This is obtained by subtracting, for each subject, the face in neutral expression, \ie, with no expression or lip movement, from each instance of the corresponding subject in both datasets. Based on this, the datasets $D_T$ and $D_E$ are represented as:
\begin{equation}
    D_T = \left \{(M_i^{t}, M_i^n)\right \}_{i=0}^P, 
    \hspace{0.3cm}
    D_E = \left \{(M_j^{e}, M_j^n)\right \}_{j=0}^H ,
\end{equation}

\noindent 
where, $M_i^{t}$ and $M_j^{e}$ represent, respectively, the $i$-th talking head and the $j$-th emotional head, while $M_i^n$, $M_j^n$ denote the corresponding neutral configurations.
$P$ and $H$ are the number of samples in the $D_T$ and $D_E$ datasets, respectively. 
For each face $M_i^{t} \in D_T$ and $M_j^{e} \in D_E$, the \textit{dense deformation offset} is computed as the difference between the animated face and its neutral configuration:
\begin{equation}
    S_i^{t} = M_i^{t} - M_i^{n}, 
    \hspace{2cm}
    S_j^{e} =  M_j^{e} - M_j^n.
\end{equation}
\noindent
This processing step generates a displacement-based representation encoding speech and expression motions from the two source datasets $S_T = \left \{S_i^{t} \right \}_{i=0}^P$, and $S_E = \left \{S_j^{e} \right \}_{j=0}^H$. 
Though both datasets contain sequences, the frames are treated individually. Thus, there is no need for the length of the sequences to match.

\subsection{Double Encoder/Shared Decoder Architecture}\label{sec:framework}
Our proposed architecture is made up of a Double-Encoder and a Shared-Decoder (DE-SD), and is summarized in~\cref{fig:overview}.
The building block for both encoders and decoder is the \textit{SpiralNet} proposed by Bouritsas~\etal~\cite{bouritsas2019neural}. 
The two distinct encoders independently process speech or expression related data: $E_T$ uses samples in $S_T$ to learn capturing speech-related movements, while $E_E$ processes samples in $S_E$ to learn deformations due to expressions. Both encoders embed the samples into separate latent vectors: $f_i^{t} = E_T(S_i^{t})$ and $f_j^{e} = E_E(S_j^{e})$.
However, these feature vectors eventually need to be combined so that the decoder can learn both motions and generate mixed deformations. Given that no ground-truth exists for the combined motions, how the features are concatenated depends on whether we are training or testing the model.

\vspace{-0.3cm}
\subsubsection{Training Phase}
Ground-truth samples are available only for each dataset separately, thus each encoder is trained alternatively. However, we need to maintain a consistent embedding size at both training and inference. Hence, the embedded feature vectors are duplicated and concatenated during training. These are then fed to the decoder to reconstruct the input: $\hat{S_i^{t}} = D(f_i^{t} \bigoplus f_i^{t})$ for $E_T$, and $\hat{S_j^{e}} = D(f_j^{e} \bigoplus f_j^{e})$ for $E_E$. This strategy allows us to later combine features from both encoders.
The training objective is to reconstruct the input displacements. We used a weighted $L_2$ loss:
\begin{equation}
L = \frac{1}{N}\sum\limits_{i=1}^{N} w_{i} \left \|S_{i} - \hat{S}_{i} \right \|_2 .
\label{eq:loss}
\end{equation}

\noindent
where $N$ represents the number of vertices in the meshes, while $S_{i}$ and $\hat{S}_{i}$ are respectively, the ground-truth and predicted displacements at vertex $i$. 
Here, $w_{i}$ represents the per-vertex weight as introduced by  Otberdout~\etal~\cite{otberdout2022sparse} that measures the contribution of each vertex in the mesh according to the inverse of its distance from the closest landmark. This factor penalizes points far from movable face areas, \ie, mouth, eyes, as they remain mostly stationary and so do not contribute to modeling facial deformations. 

\vspace{-0.4cm}
\subsubsection{Inference Phase}\label{sec:inference}
Given talking displacements $S_i^{t}$, and emotional displacements $S_j^{e}$, we use the encoders to embed them, deriving $f_i^{t} = E_T(S_i^{t})$ and $f_j^{e} = E_E(S_j^{e})$.
These feature vectors are then concatenated and fed to the decoder to generate the mixed motion: $S^{te} = D(\mu_t f_i^{t} \bigoplus \mu_e f_j^{e})$. Here, $\mu_t$ and $\mu_e$ are coefficients that allow us to adjust the contribution of the two sets of features. By modifying these coefficients, we can control the interplay between the talking and emotional displacements, significantly augmenting the span of the generation variety.
Once we obtain $S^{te}$, we can add these displacements to a neutral face and obtain the 3D emotional talking head as $M_i^{te} = M_i^n + S_i^{te}$.

\subsubsection{On the Architectural Design}\label{subsec:motivation}
Features in the two datasets are disentangled as speech and expression deformations involve \textit{different face regions}, so their spatial distribution overlaps only to some extent. Thus, to correctly reconstruct the displacements (either $S^t$ or $S^e$) during training, the decoder needs to output spatially different, almost complementary, displacement fields. Our intuition to duplicate the features is that the decoder is made up of spiral convolution layers, which operate locally, and are based on the knowledge of the mesh graph~\cite{bouritsas2019neural}. Thus each ``neuron'' in each layer corresponds to a vertex in the (sub-sampled) mesh. Since the two deformations are spatially disentangled, speech embeddings $f^t$ activate different decoder features with respect to emotion embeddings $f^e$ so that there is no need to explicitly factorize the latent space. Even if features are duplicated during training, the prior spatial disentanglement forces the decoder to activate the same set of neurons. When features are combined, it instead induces the activation of all neurons, leading to realistic entanglement of the deformation (see~\cref{fig:actmaps}).

\section{Experimental Results}\label{sec:experiments}
In the following, we first introduce the datasets and the metrics (\cref{sec:datasets-metrics}), then we summarize the methods for emotional talking heads generation that we trained on the data synthesized with our DE-SD architecture (\cref{sec:baselines}). 
We report quantitative and qualitative results plus a test involving the proposed methods in~\cref{sec:quantitative}, and~\cref{sec:baselines-evaluation}, respectively. Finally, in~\cref{sec:user}, we conducted two user studies to compare our approach with current SOTA methods for emotional talking heads, \textit{i.e.}, Emotalk~\cite{peng2023emotalk} and EMOTE~\cite{emote}.

\subsection{Datasets and Metrics}\label{sec:datasets-metrics}
We utilized two datasets, VOCAset and Florence~4D, to train our DE-SD network, which was subsequently used to generate the EmoVOCA dataset.

\noindent
\textbf{VOCAset}~\cite{Cudeiro_2019_CVPR} contains 3D talking head sequences from 12 actors, evenly split between 6 males and 6 females. Each actor recorded 40 sentences, with durations ranging from 3 to 5 seconds. The dataset provides per-frame 3D facial reconstructions captured at 60 frames per second (fps) along with corresponding audio recordings. However, VOCAset lacks head movements and upper facial expressions.\\
\textbf{Florence 4D}~\cite{10042606} offers 3D dynamic sequences representing 70 different emotions and facial expressions. Each sequence consists of 60 meshes, with the peak facial expression typically occurring between frames 25 and 35.\\
\textbf{EmoVOCAv1} is a collection of expressive 3D talking head sequences generated using our DE-SD architecture. It combines features from VOCAset and Florence 4D to generate sequences for five emotions: \textit{afraid}, \textit{angry}, \textit{disgust}, \textit{happy}, and \textit{sad}, each with three intensity levels by varying the $\mu_e$ factor. This produced a total of 7,200 3D sequences.\\
\textbf{EmoVOCAv2} expands upon EmoVOCAv1, adding 
\textit{moody}, \textit{drunk}, \textit{ill}, \textit{suspicious}, \textit{pleased}, and \textit{upset}, for a total of 11 emotions, with three intensity levels, resulting in 15,840 sequences. All datasets were divided into training (8 actors), validation, and test sets (2 actors each).

Following previous works~\cite{FaceDiffuser_Stan_MIG2023, fan2022faceformer, peng2023emotalk}, we used the following metrics, measured in millimeters:
\begin{itemize}
    \item \textbf{MVE} (Max Vertex Error): The maximum $L_2$ error between the predicted and ground truth mesh vertices, providing a global measure of reconstruction accuracy.
    \item \textbf{UVE} (Upper Vertex Error): The maximum $L_2$ error computed on the vertices in the eyes and forehead regions, focusing on emotion-related movements.
    \item \textbf{LVE} (Lip Vertex Error): The maximum $L_2$ error computed on the vertices around the mouth region, capturing both speech-/emotion-related movements.
\end{itemize}

\noindent
MVE was used as a global metric to evaluate the overall reconstruction quality of DE-SD, as shown in~\cref{tab:quantitative}. For talking head evaluation, UVE and LVE were chosen as local descriptors to assess the quality of expression-related and speech-related movements, respectively, as shown in~\cref{tab:evaluation}.

\subsection{Emotional 3D Talking Head Generators}\label{sec:baselines}
We used both EmoVOCAv1 and EmoVOCAv2 to train two properly customized state-of-the-art solutions for 3D talking heads generation, namely Faceformer~\cite{fan2022faceformer}, and S2L+S2D~\cite{nocentini2023learning}. 
These methods are constrained to generate neutral talking heads, so we adapted them for training and testing on our data. The primary goal is to show that by training on EmoVOCAv1/v2 we can both \textit{(i)} surpass the accuracy of SOTA methods trained on the original VOCAset, and \textit{(ii)} perform better than state-of-the-art emotional talking head generators that exploit 2D video data and parametric head models.  
Both the models were conditioned using audio features extracted from the wav2vec 2.0~\cite{wav2vec} encoder, which was shown to be highly effective~\cite{fan2022faceformer, Thambiraja_2023_ICCV, nocentini2023learning, peng2023emotalk, emote}. 
To add the emotion information and condition the generation, we represented each emotion label $L$, and intensity label $I$, as a one-hot vector, and embedded them as $64$-dimensional vectors using a linear layer.\\
\textbf{Emo(E)-Faceformer} 
is inspired by Faceformer~\cite{fan2022faceformer}. The architecture proposed by Fan~\etal involves a transformer decoder, trained autoregressively, conditioned with audio features. We modified the original architecture by concatenating the two embeddings $L$, $I$ resulting from the emotion and intensity labels. Similar to Faceformer, we trained the model autoregressively with the same masking and the same periodic positional encoding. 
This method learns to animate a 3D face given an audio input, an emotion label and an intensity label. The $L_2$ loss between the prediction and the ground truth was used for training.\\
\textbf{Emo(E)-S2L+S2D}
develops upon the S2L+S2D method  in~\cite{nocentini2023learning}. This method works by decoupling the generation in two steps: first, a bi-directional LSTM model (S2L) generates the motion of 68 landmarks of the face from the audio signal. Then, a mesh decoder (S2D) expands the landmarks motion to all the vertices of the face mesh. We trained the E-S2L with the landmarks extracted from the meshes, then we separately trained the S2D decoder to densify the movement of landmarks into the face mesh vertices. Similar to E-FaceFormer, we concatenated the two embeddings $L$, $I$ to the audio features. The E-S2L learns to predict landmarks movement based on an audio input, an emotion label, and an intensity label. To train the E-S2L model, we used the average $L_2$ loss between the prediction and the ground truth. To train the S2D, we used the weighted $L_2$ defined in~\eqref{eq:loss}.

\subsection{EmoVOCA Evaluation}\label{sec:quantitative}
\noindent
\textbf{Quantitative results.} We first assess the accuracy of our DE-SD model to reconstruct the test sets of VOCAset and Florence~4D. In~\cref{tab:quantitative}, we report the results of an ablation study aimed at evaluating the effect of the weighted $L_2$ loss applied during training and a different design for DE-SD trained with a single, shared encoder. 

\vspace{-0.2cm}
\begin{table}[!ht]
\centering
\caption{Ablation study. Contributions of: weighted loss in~\cite{otberdout2022sparse} (left); double or single encoder (right). Errors are in mm.}
\resizebox{0.99\linewidth}{!}{
\begin{tabular}{l|c||c}
\toprule
& \multicolumn{1}{c}{\textbf{VOCAset}} & \multicolumn{1}{c}{\textbf{Florence 4D}} \\
\midrule
\hline
\textbf{Loss} & \textbf{MVE} $\downarrow$ & \textbf{MVE} $\downarrow$ \\
\midrule
Standard $L_2$ & 0.931 & 0.844\\
\hline           
\textbf{Weighted $L_2$} & \textbf{0.722} & \textbf{0.657}\\
\midrule
\end{tabular}
\newline
\hspace*{1cm}
\newline
\begin{tabular}{l|c||c}
\toprule
& \multicolumn{1}{c}{\textbf{VOCAset}} & \multicolumn{1}{c}{\textbf{Florence 4D}} \\
\midrule
\hline
\textbf{Encoder} & \textbf{MVE} $\downarrow$ & \textbf{MVE} $\downarrow$ \\
\midrule
Single & 0.745  & 0.676\\
\hline
\textbf{Double} & \textbf{0.722} & \textbf{0.657}\\
\midrule
\end{tabular}
}
\label{tab:quantitative}
\end{table}

\vspace{-0.3cm}
\begin{figure*}[!ht]
    \centering
    \begin{subfigure}[t]{0.65\textwidth}
        \centering
        \includegraphics[width=\textwidth]{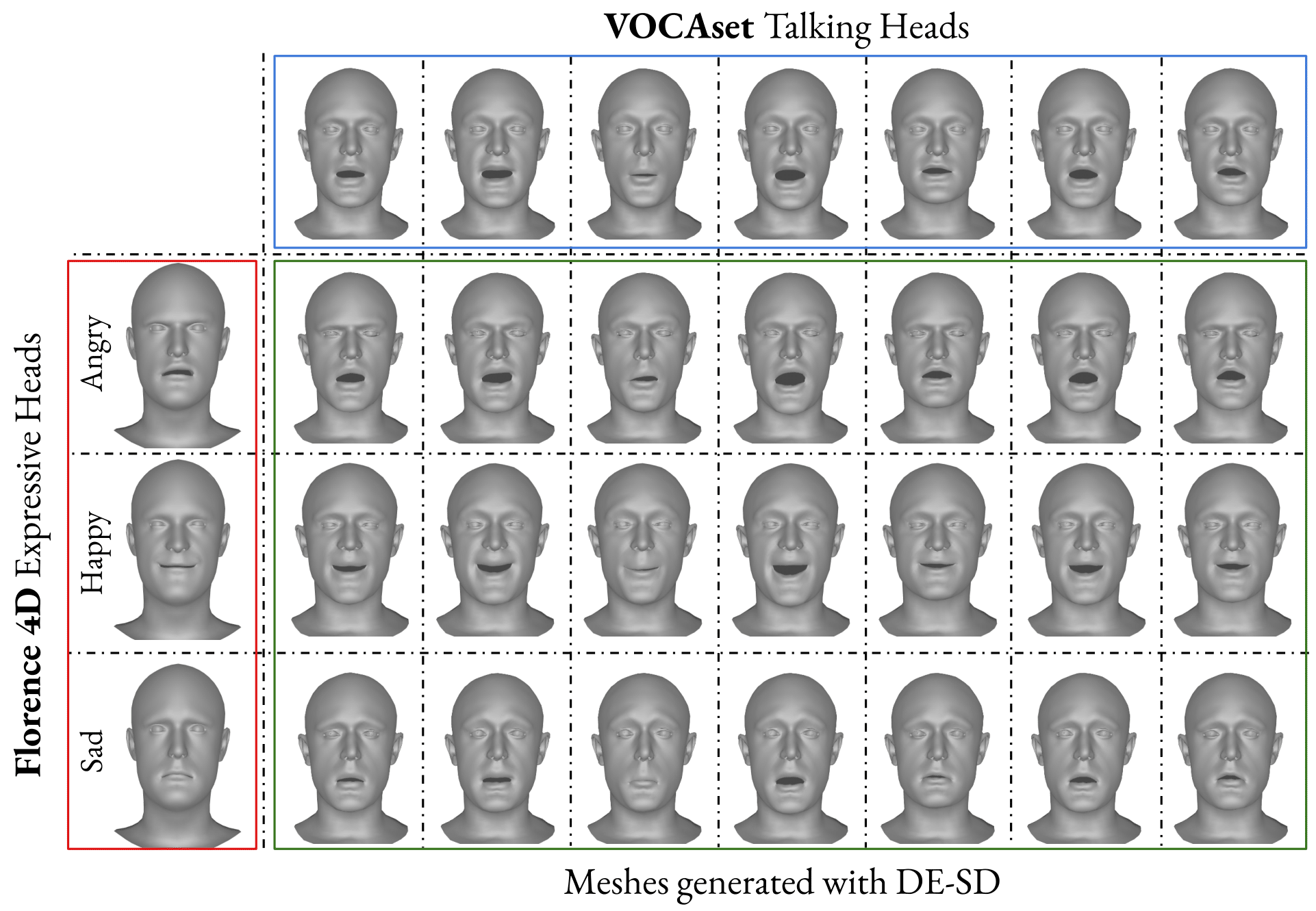}
        \caption{DE-SD Qualitative Results. Meshes with speech combined with expressions are shown.}
        \label{fig:qualitative}
    \end{subfigure}%
    \hfill
    \begin{minipage}[t]{0.35\textwidth}
        \vspace{-8cm}
        \centering
        \begin{subfigure}[t]{0.6\textwidth}
            \centering
            \includegraphics[width=\textwidth]{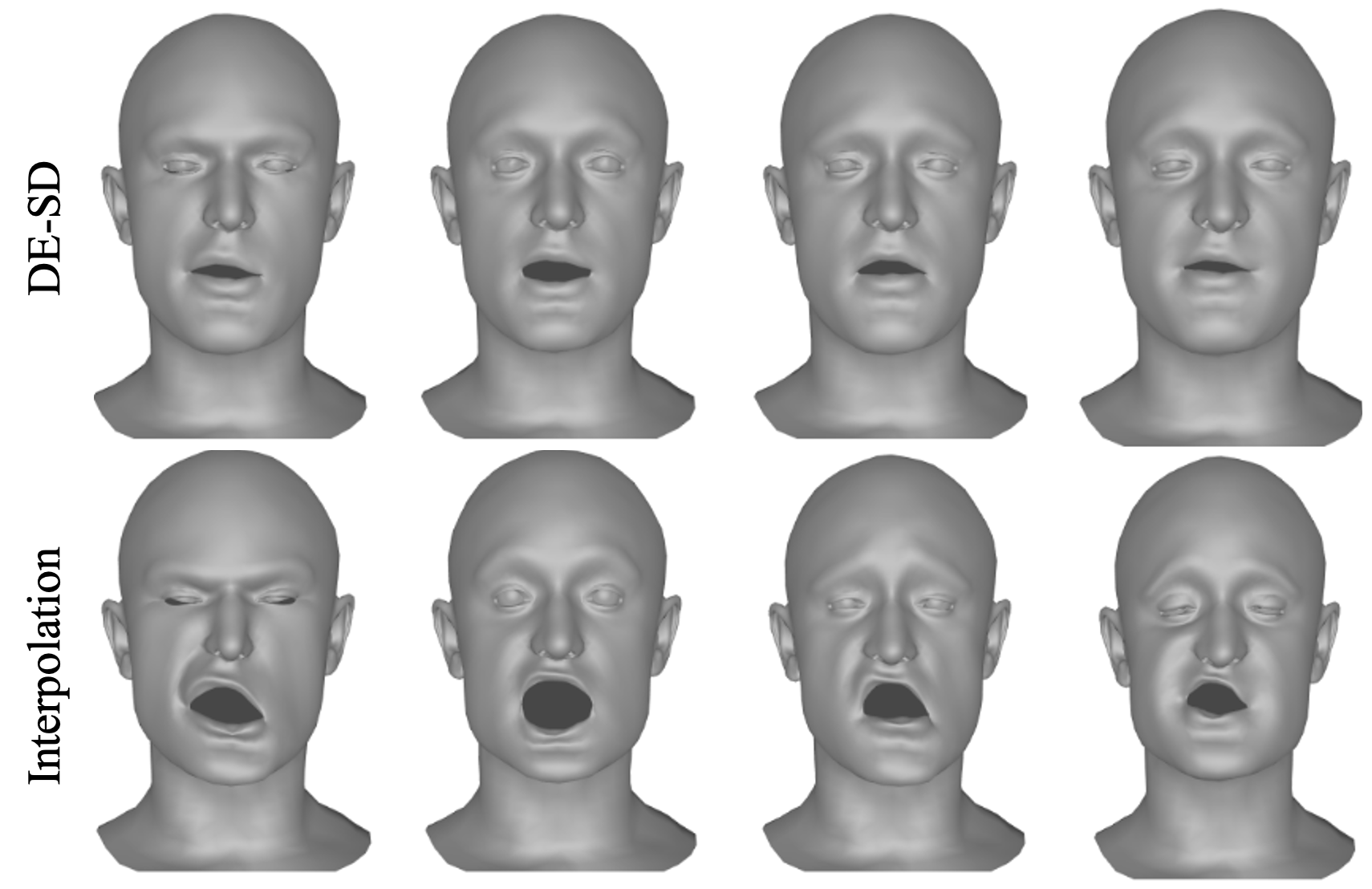}
            \caption{Interpolation vs. DE-SD}
            \label{fig:interp}
        \end{subfigure}\\[1em] 
        \begin{subfigure}[t]{\textwidth}
            \centering
            \includegraphics[width=\textwidth]{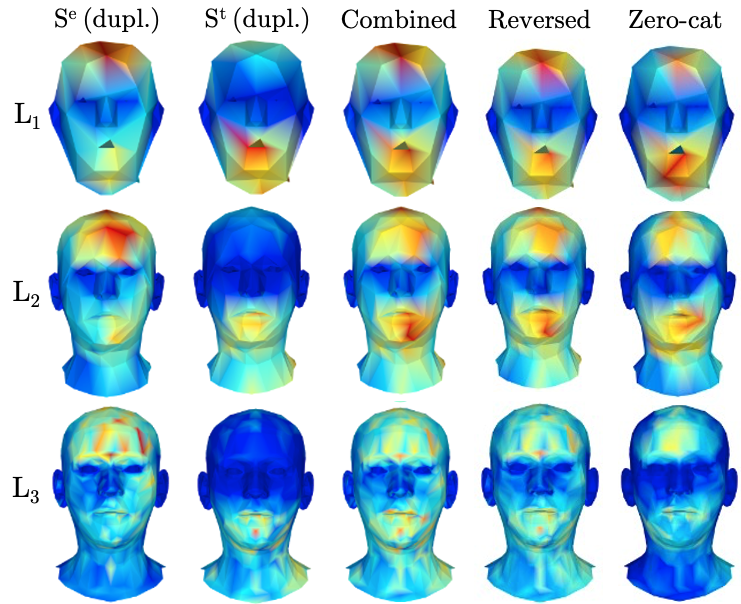}
            \caption{Decoder features in layers $L_i$. Red/blue=high/low.}
            \label{fig:actmaps}
        \end{subfigure}
    \end{minipage}
    \caption{\textbf{(a)} The blue box (top row) shows some meshes from VOCAset, while the red box (left column) showcases various expressive heads from Florence 4D. In the green box, the face at row $i$ and column $j$ is the result of the combination of the $i$-th expressive and $j$-th talking heads, \ie, $S^{te} = D(E_T(S_j^t) \bigoplus E_E(S_i^e))$. So, each column includes the talking head on the top conditioned with different expressions. \textbf{(b)} Comparison of DE-SD (top row) with displacement interpolation (bottom row). Simple interpolation in 3D space leads to unrealistic deformations and loss of speech accuracy. 
    \textbf{(c)} Comparison of several strategies of feature combination. The decoder features are shown for each layer $L_i$. Even if features are duplicated during training, the prior spatial disentanglement forces the decoder to activate different neurons. When combined, the features induce a combined action of all neurons, leading to realistic entanglement. }
    \vspace{-0.2cm}
    \label{fig:features}
\end{figure*}

We first investigate the effect of the weighted loss~\cite{otberdout2022sparse}. Predictably, the latter leads to significantly improved outcomes as the model better focuses on movable facial areas. 
Another significant result is the improvement achieved by employing a double-encoder. This outcome aligns with our expectations, considering that each encoder specializes in extracting facial features that are either speech or emotion-related, further pushing a disentangled feature representation. This specialization arises from the datasets difference: in VOCAset, the upper face remains stable, whereas in Florence 4D it plays a more pivotal role (Sect.~\ref{subsec:motivation}). 

To further assess the feature concatenation strategy of DE-SD, we trained a simple expression classifier (Spiral Convolution Encoder with an MLP Head) using the Florence4D mesh sequences (expression-only). We then tested it on five different versions of our EmoVOCA dataset (expression+speech) generated using five different feature combination strategies in DE-SD. This was done to evaluate how well expressions are preserved when combined with speech. We trained DE-SD using different strategies: \textit{(i)} summing (Sum) or multiplying (Mult) the embeddings $f^t$ and $f^e$; \textit{(ii)} concatenating neutral (zero) vectors instead of duplicating the embeddings; and \textit{(iii)} testing the original DE-SD by reversing the positions of $f^t$ and $f^e$ to verify the intuitions discussed in~\cref{subsec:motivation}. The results for each intensity level are shown in~\cref{tab:classificator}. We also compared the geometric similarity of the generated meshes against those in the original datasets to determine which method best preserves facial features. Specifically, we utilized the LVE for VOCAset  and the UVE for Florence 4D. Both complementary measures prove that duplicating features during DE-SD training leads to more recognizable expressions, even when the embeddings were reversed at inference, outperforming the other strategies. 

\vspace{-0.1cm}
\begin{table}[!ht]
\centering
\caption{Classification accuracy for different expr. intensities. Error on both training datasets.}
\vspace{-0.2cm}
\resizebox{0.99\linewidth}{!}{
\begin{tabular}{l|ccc|cc}
\toprule
\textbf{Model} & $I_1$ Acc & $I_2$ Acc & $I_3$ Acc & \textbf{LVE VOCAset} $\downarrow$  & \textbf{UVE Florence 4D} $\downarrow$ \\
\midrule
DE-SD Sum & 0.40 & 0.64 & 0.72 & 6.181 & 1.073\\
DE-SD Mult & 0.33 & 0.55 & 0.66 & 6.897 & 1.248\\
DE-SD Zero-cat & 0.37 & 0.58 & 0.69 & 6.324 & 1.137\\
\hline
DE-SD Reversed & 0.45 & 0.68 & 0.76 & 5.872 & \textbf{0.940}\\
DE-SD & \textbf{0.46} & \textbf{0.71} & \textbf{0.82} & \textbf{5.861} & 0.981\\
\midrule
\end{tabular}
}
\label{tab:classificator}
\vspace{-0.2cm}
\end{table} 
\noindent
\textbf{Qualitative Results.} \cref{fig:qualitative} shows various expressive meshes combined with talking heads with DE-SD. The deformation of the upper face is particularly influenced by the expressive mesh, although the mouth shape is influenced by the expression as well, even if to a lesser extent. This unbalanced influence is valuable as it infuses expressiveness to the talking heads, at the same time preserving the lip movements that correspond to the spoken sentence. Using two separate encoders led to an effective disentangled learning of the motion priors, leading to a strong compositional ability~\cite{shi2021SemanticStyleGAN}. This is instead compromised in previous methods such as EmoTalk or EMOTE (see~\cref{sec:user}). 

It could be argued that the same goal could be achieved by simply linearly combining the speech and expression offsets ($S_i^t$ and $S_i^e$) or by morphing the faces using some Morphable Model parameters. As depicted in~\cref{fig:interp}, this is not the case. Displacement interpolation introduces severe artifacts, and leads to the loss of speech-related movements, resulting in lower realism in the generated faces.
By directly summing the expressive offsets $S_i^e$ to inexpressive talking heads generated by FaceFormer or S2L-S2D, we also quantitatively show that this is not a viable solution either (see results in~\cref{tab:evaluation-emovoca}).
Finally, examples in~\cref{fig:actmaps} qualitatively support the technical choice expounded in Sect.~\ref{subsec:motivation} and the results in~\cref{tab:classificator}. The average activation of decoder features are shown for each layer $L_i$. Duplicating either $f^t$ or $f^e$ activate distinct decoder features and, when combined, all features activate, contributing to a realistic entanglement. Reversing their arrangement does not have a significant influence, meaning that the two encoders effectively captured the disentangled deformations. Forcing an explicit latent arrangement by concatenating a zero vector would lose this ability, also resulting in samples of poorer quality.


\subsection{Generating Emotional 3D Talking Heads}\label{sec:baselines-evaluation}
Both E-Faceformer and E-S2L-S2D models were trained using separately both EmoVOCAv1 and EmoVOCAv2. 
We report average generation results across the test set of both versions in~\cref{tab:evaluation-emovoca}. 
Based on the results presented in~\cref{tab:evaluation}, it is evident that training a model using EmoVOCAv2 yields improved performance compared to training with EmoVOCAv1. This enhancement is a result of the increased number of samples in EmoVOCAv2.
The E-S2L-S2D method clearly performs favorably with respect to E-FaceFormer (consistently with the result reported in~\cite{nocentini2023learning}). 

To conduct a more comprehensive evaluation of our generators E-Faceformer and E-S2L+S2D, we performed additional tests using the (inexpressive) VOCA-Test as suggested by~\cite{peng2023emotalk}. We set the one-hot encoding of both emotion and intensity to zero, for both E-Faceformer and E-S2L+S2D. Results in~\cref{tab:evaluation-zero-shot} show that our emotional talking head generators exhibit good generalization ability to speech-related movements, surpassing the performance of the same architectures trained on the original VOCAset. Moreover, E-S2L+S2D yields superior performance compared to both EmoTalk and EMOTE. These outcomes further highlight the good quality of the samples synthesized with our DE-SD, suggesting our solution as a viable way to create complex yet effective 3D datasets.

\begin{table}[!ht]
\centering
\caption{Quantitative comparison with previous works: (a) test on EmoVOCAv1 and EmoVOCAv2; (b) tests on VOCA-Test. (v1) and (v2) indicate training on EmoVOCAv1 and EmoVOCAv2, respectively; (Voca) indicates training on VOCAset; (2D) indicates training on 3D datasets reconstructed from videos.}
\vspace{-0.3cm}
\resizebox{\linewidth}{!}{
\begin{subtable}[t]{0.44\textwidth}
\begin{tabular}{l||c||c}
\hline
\textbf{Baseline} & \textbf{LVE} (mm) $\downarrow$ & \textbf{UVE} (mm)  $\downarrow$\\
\hline 
\hline
 Faceformer + $S^e$ & 5.971 & 1.923 \\
 S2L+S2D + $S^e$ & 4.872 & 1.467 \\
 \hline
 E-Faceformer(v1) & 3.425 & 0.927 \\
 E-S2L+S2D(v1)  & 2.165 & 0.552 \\
 \hline
 E-Faceformer(v2) & 3.134 & 0.904 \\
 E-S2L+S2D(v2)  & \textbf{1.845} & \textbf{0.501} \\
 \hline
\end{tabular}
\caption{\large Results from the baselines. The first two rows are tested on EmoVOCAv1.}
\label{tab:evaluation-emovoca}
\end{subtable}
\hspace*{0.2cm}
\begin{subtable}[t]{0.44\textwidth}
\begin{tabular}{l||c||c}
\hline
\textbf{Method} & \textbf{LVE} (mm) $\downarrow$  & \textbf{UVE} (mm) $\downarrow$ \\ 
\hline 
\hline
 VOCA(Voca) & 6.993 & 1.201\\ 
 Faceformer(Voca) & 6.123 & 1.117\\ 
 SelfTalk(Voca) & 5.618 & 0.989\\ 
 S2L+S2D(Voca)  & 4.789 & 0.934\\ 
 FaceDiffuser(Voca)  & 4.350 & 0.925\\
 CodeTalker(Voca)  & 3.651 & 0.912\\
 \hline
 EmoTalk(2D) & 4.134* & - \\ 
 EMOTE(2D) & 4.561 & 0.897\\
 \hline 
 E-Faceformer(v1) & 4.371 & 0.964\\ 
 E-S2L+S2D(v1)  & 3.697 & 0.867\\ 
 \hline 
 E-Faceformer(v2) & 3.789 & 0.912\\ 
 E-S2L+S2D(v2)  & \textbf{3.181} & \textbf{0.791}\\ 
 \hline
\end{tabular}
\caption{\large Results on VOCA-Test dataset. For the EmoTalk (*) model, we report the results as presented in the original paper.}
\label{tab:evaluation-zero-shot}
\end{subtable}
}
\vspace{-0.3cm}
\label{tab:evaluation}
\end{table}

\begin{figure}[!ht]
\centering
    \begin{subfigure}[b]{0.23\linewidth} 
        \centering
        \includegraphics[width=\linewidth]{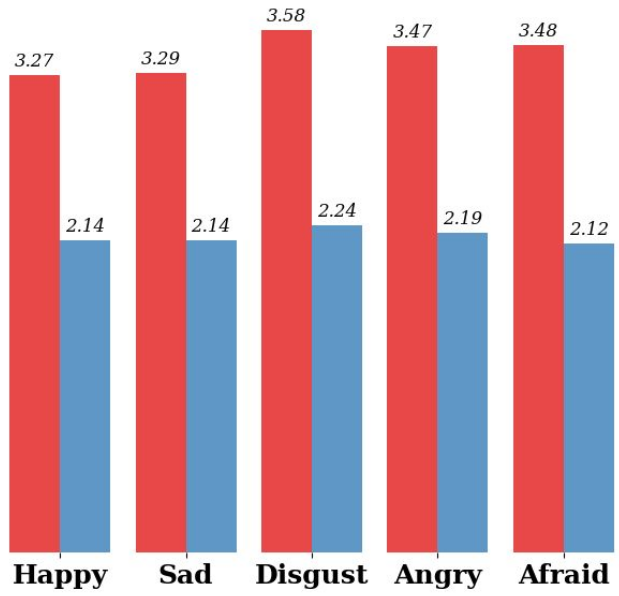}
        \caption{\tiny LVE Emotion}
        \label{fig:lve-emo-comparison}
    \end{subfigure}%
    \hspace{0.1cm}
    \begin{subfigure}[b]{0.23\linewidth} 
        \centering
        \includegraphics[width=\linewidth]{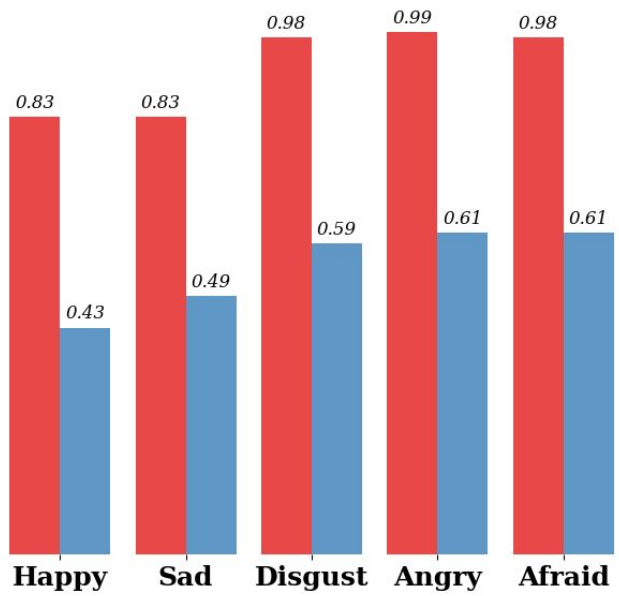}
        \caption{\tiny UVE Emotions}
        \label{fig:uve-emo-comparison}
    \end{subfigure}
    \begin{subfigure}[b]{0.24\linewidth} 
        \centering
        \includegraphics[width=\linewidth]{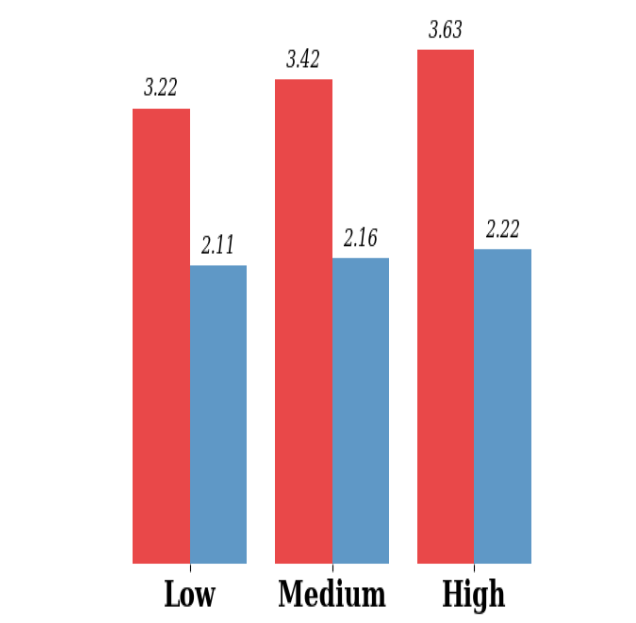}
        \caption{\tiny LVE Intensities}
        \label{fig:lve-int-comparison}
    \end{subfigure}
    \begin{subfigure}[b]{0.24\linewidth} 
        \centering
        \includegraphics[width=\linewidth]{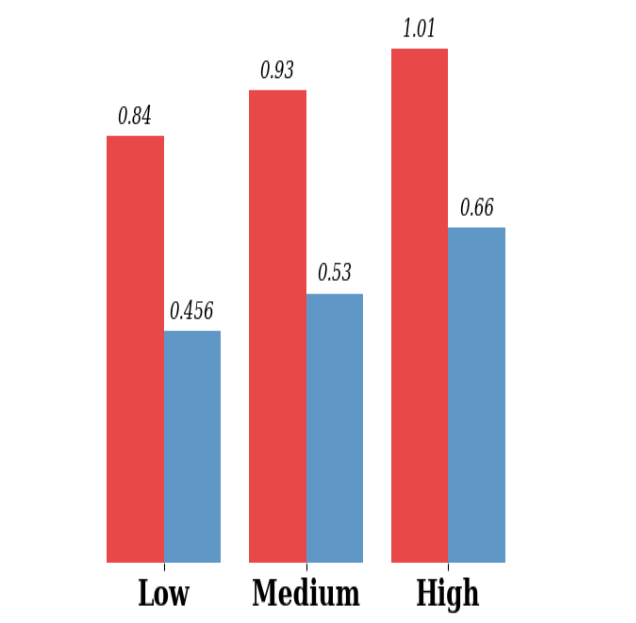}
        \caption{\tiny UVE Intensities}
        \label{fig:uve-int-comparison}
    \end{subfigure}
    \caption{Generation results comparison for each emotion and each intensity level. The blue bars refer to E-S2L-S2D, while red bars to E-FaceFormer. Both baselines are trained and tested on EmoVOCAv1.}
    \label{fig:comparison-bars}
    \vspace{-0.4cm}
\end{figure}

For a more detailed analysis, in~\cref{fig:comparison-bars} we report results for all the emotions and intensity levels separately. We observe a rather stable accuracy across the different emotions and intensities, although the UVE measure grows for those involving stronger movements of the upper face, such as \textit{angry} or intensity \textit{high}. 
In~\cref{fig:qualitative_gen}, we show qualitative examples also for EmoTalk and EMOTE (quantitative comparison was not possible, see~\cref{sec:user}). Note that our samples and the EMOTE ones are generated performing a zero-shot test using an audio from RAVDESS~\cite{livingstone2018ryerson}. This was needed because EmoTalk only infers the emotion from the audio, and the VOCAset includes only neutral speech. 
Qualitatively, the results for E-S2L-S2D are the most realistic both in terms of speech mimicking and perceived emotion (see also~\cref{tab:user} and~\cref{tab:user-emote}). Notably, EmoTalk can convey some emotional content, yet only by slightly moving the upper face area, whereas the mouth region remains rather static. This results both in smaller emotional effects, and reduced speech realism. EMOTE, instead, can generate good and recognizable emotional content, yet it is less accurate in terms of lip synchronization (see~\cref{tab:user-emote}).

\vspace{-0.1cm}

\subsection{E-S2L+S2D Additional Features}\label{sec:addfeat}
\begin{figure*}[!ht]
\centering
    \begin{minipage}[b]{0.55\textwidth} 
    \vspace{-2cm}
        \centering
        \includegraphics[width=\textwidth]{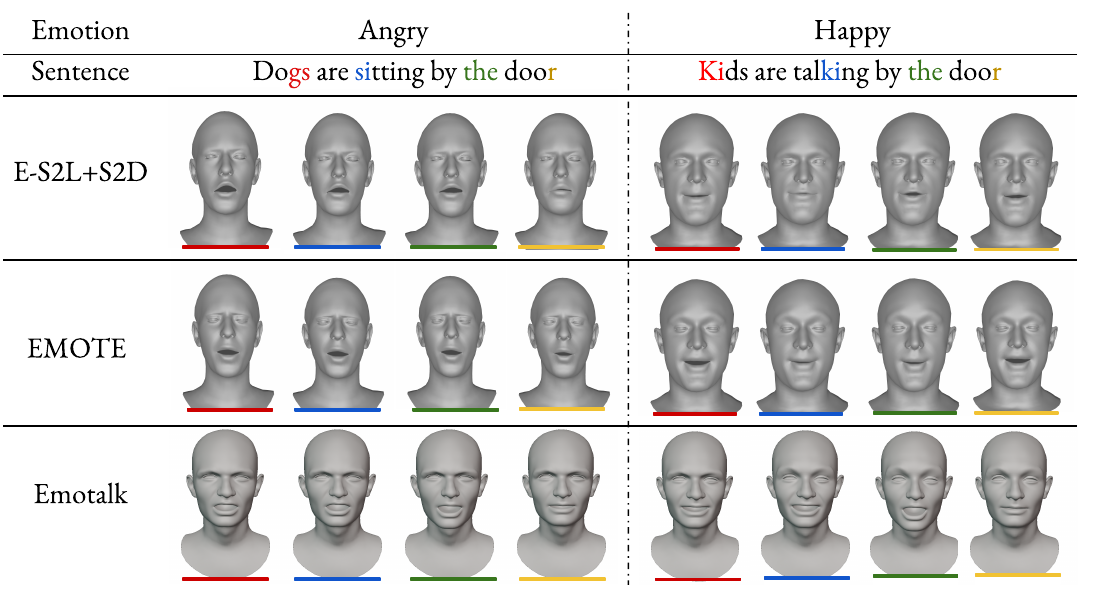} 
        \caption{Qualitative examples generated by E-S2L+S2D trained on EmoVOCAv1 in comparison with EmoTalk and EMOTE. Colors represent specific moments in time.}
        \label{fig:qualitative_gen}
    \end{minipage}%
    \hfill
    \begin{minipage}[b]{0.43\textwidth} 
        \centering
        \includegraphics[width=\textwidth]{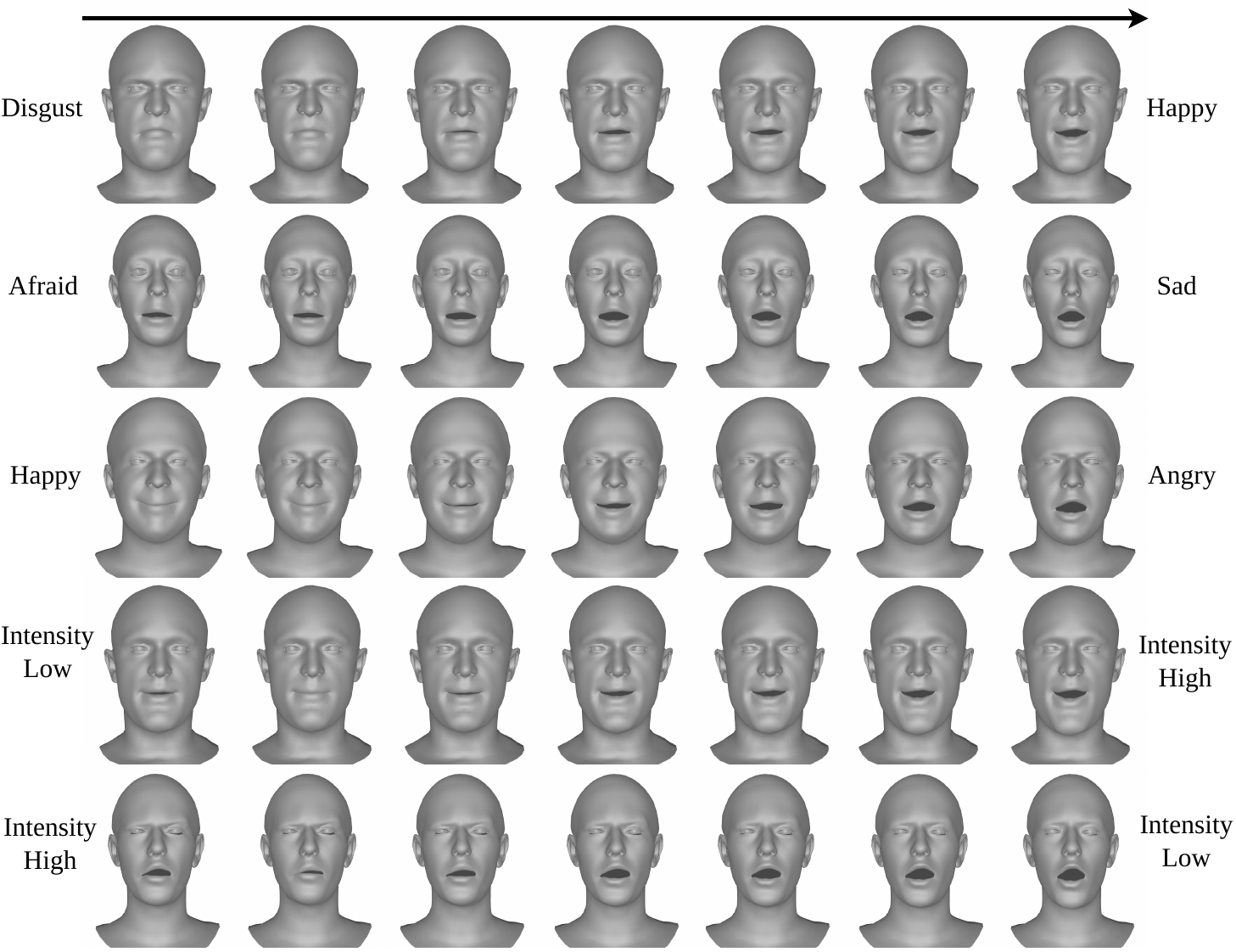} 
        \caption{Additional feature of E-S2L+S2D: varying the one-hot labels during inference, we are able to change the emotion or intensity displayed by the talking head.}
        \label{fig:add_features}
    \end{minipage}
    \vspace{-0.2cm}
\end{figure*}

The E-S2L+S2D model, trained with EmoVOCA, efficiently generates emotional talking heads based on one-hot encoding of emotion and intensity labels. While training activates only one emotion and one intensity level at a time, the model's high generalization enables diverse capabilities during inference. It can produce sequences with temporal emotion or intensity transitions, varying emotion and intensity labels during generation. In~\cref{fig:add_features}, we can see several examples of emotion and intensity transition by varying the one-hot encodings in a chosen moment in time. 

These capabilities emerge from both the model's design and the EmoVOCA dataset used for training. Consequently, a model trained on EmoVOCA can generate sequence of faces with emotion and intensity variation despite being trained on separate tasks and not encountering such scenarios during training. We highlight this a unique feature that is not shown by any other method or dataset. Qualitative examples can be found in the supplementary material.

\subsection{User based Study}\label{sec:user}
To our knowledge, EmoTalk~\cite{peng2023emotalk} and EMOTE~\cite{emote} are the only approaches in recent literature for generating emotional 3D talking heads.
Quantitative comparison is not possible due to discrepancies in mesh topology and dataset utilization across EmoTalk and EMOTE. EmoTalk's mesh topology diverges from FLAME, while EMOTE's training dataset differs from ours, impeding direct evaluation.
Our decision to not re-implementing EmoTalk is based on its similarity to FaceFormer, employing a transformer decoder, with emotion inferred from audio. Similarly, retraining EMOTE with our dataset was not feasible due to its reliance on 2D video-based loss computation, which we lack. Consequently, we conducted two user-based studies involving 22 and 20 participants, spanning expert and non-expert, to assess our E-S2L+S2D trained on EmoVOCAv1 in comparison with EmoTalk and EMOTE.
Following methodologies in prior works~\cite{peng2023emotalk,fan2022faceformer,Thambiraja_2023_ICCV}, we designed A/B studies for user evaluation. Thirty videos, generated using E-S2L+S2D trained on EmoVOCAv1, EmoTalk and EMOTE, were divided as follows:
\emph{(i)} Ten videos generated from RAVDESS~\cite{livingstone2018ryerson} test set audios, asking users to assess lip synchronization fidelity and emotion portrayal (Test~1);
\emph{(ii)} Ten videos generated from \textit{in-the-wild} YouTube audios, asking evaluations on lip synchronization (Test~2);
\emph{(iii)} Ten videos from RAVDESS audios were presented to users \textit{without audio}, asking them to classify the emotion the generated videos seem to portray, so to verify if users can infer the emotion looking at the animation (Test~3).

Results from both user studies, summarized in~\cref{tab:user} and~\cref{tab:user-emote}, clearly favors E-S2L+S2D trained on EmoVOCAv1 for its superior speech realism and emotional expression. Tables report users preference on the left, and classification accuracy in the confusion matrices. Notably, despite EmoTalk and EMOTE being trained on 10K+ sentences compared to our 40, our model demonstrates enhanced generalization to in-the-wild audios (Test~2). A subset of the videos utilized in both user studies can be accessed in the supplementary material. 

\begin{table}[!ht]
\centering
\caption{User study comparing E-S2L-S2D and EmoTalk. \textbf{Left}: Test~1 and Test~2. \textbf{Right}: confusion matrix of Test~3 (in \%, ``Ours/EmoTalk'', for \textit{Happy}, \textit{Disgust}, \textit{Fear}, \textit{Sad}, \textit{Angry}).}
\vspace{-0.1cm}
\resizebox{0.99\linewidth}{!}{
\begin{tabular}{l|c|c}
\toprule 
Test & Criterion & Ours vs. EmoTalk~\cite{peng2023emotalk} \\
\toprule
\multirow{2}{*}{Test~1} & Lip Sync & \textbf{72.7}\% / 27.3\% \\
                & Emotion & \textbf{66.6}\% / 33.4\%\\
\midrule       
\multirow{1}{*}{Test~2} & Lip Sync & \textbf{76.2}\% / 23.8\% \\
\bottomrule                        
\end{tabular}
\newline
\hspace*{1cm}
\newline
\begin{tabular}{c|c|c|c|c|c}
\toprule 
& HA & DI & FE & SA & AN \\
\hline
HA  & \hl{\textbf{100}/95.2}   & 0.0/0.0 & 0.0/0.0 & 0.0/4.8 & 0.0/0.0 \\
DI  & 0.0/14.3 & \hl{19.0/\textbf{38.1}} & 4.8/4.8 & 42.9/9.5 & 33.3/33.3 \\
FE  & 0.0/0.0 & 5.3/14.3 & \hl{\textbf{78.9}/28.6} & 10.5/23.8 & 5.3/33.3 \\
SA  & 0.0/61.9 & 9.5/14.3 & 23.8/4.8 & \hl{\textbf{52.3}/9.5} & 14.3/9.5 \\
AN  & 0.0/0.0 & 23.8/19.0 &  4.8/4.8 & 0.0/14.3 & \hl{\textbf{71.4}/61.9} \\
\bottomrule
\end{tabular}
}
\label{tab:user}
\end{table}
\vspace{-0.2cm}

\begin{table}[!ht]
\centering
\caption{User study comparing E-S2L-S2D and EMOTE. \textbf{Left}: Test~1 and Test~2. \textbf{Right}: confusion matrix of Test~3 (in \%, ``Ours/EMOTE'', for \textit{Happy}, \textit{Disgust}, \textit{Fear}, \textit{Sad}, \textit{Angry}). }
\vspace{-0.1cm}
\resizebox{0.99\linewidth}{!}{
\begin{tabular}{c|c|c}
\toprule 
Test & Criterion & Ours vs EMOTE~\cite{emote} \\
\toprule
\multirow{2}{*}{Test 1} & Lip Sync & \textbf{82.4}\% / 17.6\% \\
                        & Emotion & \textbf{55.5}\% / 45.5\%\\
\midrule       
\multirow{1}{*}{Test 2} & Lip Sync & \textbf{83.2}\% / 16.8\%\\
\bottomrule                        
\end{tabular}
\newline
\hspace*{1cm}
\newline
\begin{tabular}{c|c|c|c|c|c}
\toprule 
& HA & DI & FE & SA & AN \\
\hline
HA  & \hl{\textbf{100}/\textbf{100}}   & 0.0/0.0 & 0.0/0.0 & 0.0/4.8 & 0.0/0.0 \\
DI  & 0.0/0.0 & \hl{\textbf{59.1}/40.4} & 0.0/0.0 & 40.9/0.0 & 0.0/59.6 \\
FE  & 0.0/11.0 & 0.0/0.0 & \hl{\textbf{89.8}/78.5} & 0.0/10.5 & 10.2/0.0 \\
SA  & 0.0/0.0 & 0.0/0.0 & 9.4/9.5 & \hl{78.4/\textbf{90.5}} & 12.2/0.0 \\
AN  & 0.0/0.0 & 24.3/0.0 &  0.0/0.0 & 0.0/0.0 & \hl{75.7/\textbf{100}} \\
\bottomrule
\end{tabular}
}
\label{tab:user-emote}
\end{table}

\vspace{-0.2cm}
\section{Conclusions and Limitations}\label{sec:conclusions}
In this paper, we introduced an innovative approach for creating 3D emotional talking heads leveraging only 3D data. Our approach demonstrated strong performance in generating expressive talking heads with high fidelity. 
Initially, we addressed the scarcity of available datasets by devising a method to combine talking and expressive unpaired 3D faces to generate a new dataset, named EmoVOCA. 
We defined two versions of this dataset with different sets of emotions. However, our DE-SD framework enables the generation of numerous additional versions by varying the selection of emotional faces used to condition the inexpressive faces in the talking head dataset.
Subsequently, we properly customized two state-of-the-art methods, E-Faceformer and E-S2L+S2D. We effectively accomplish the goal by producing more realistic emotional 3D talking heads, if compared to methods that utilize 2D videos as replacement for the lack of 3D data. Our results demonstrate convincing lip syncing and perceivable emotional content.

The proposed approach still faces some potential limitations that can be addressed as future work. 
First, as shown in the confusion matrices of~\cref{tab:user} and~\cref{tab:user-emote}, some emotions resulted more difficult to inject with our solution. 
In addition, the proposed framework still lacks some realism, like for eyes blinking, head pose, \etc. How to do so also constitutes an interesting direction to investigate. 


\bibliographystyle{plain}
\bibliography{egbib.bib}

\end{document}